\documentclass[10pt,twocolumn,letterpaper]{article}

\usepackage{cvpr}
\usepackage{times}
\usepackage{epsfig}
\usepackage{graphicx}
\usepackage{multirow}
\usepackage{amsmath}
\usepackage{amssymb}
\usepackage{color}
\usepackage{bm}
\usepackage{booktabs} 
\usepackage{array}  %

\usepackage[pagebackref=true,breaklinks=true,colorlinks,bookmarks=false]{hyperref}
\hypersetup{citecolor=blue}

\cvprfinalcopy 

\ifcvprfinal\fi
\begin{document}

\title{Proposal Learning for Semi-Supervised Object Detection}

\author{Peng Tang$^{\dag}$ \ \ \ Chetan Ramaiah$^{\dag}$ \ \ \ Yan Wang$^{\ddag}$ \ \ \ Ran Xu$^{\dag}$ \ \ \ Caiming Xiong$^{\dag}$ \\
$^{\dag}$Salesforce Research \ \ \ \ \ \ $^{\ddag}$The Johns Hopkins University\\
{\tt\small \{peng.tang,cramaiah,ran.xu,cxiong\}@salesforce.com \ \ \ wyanny.9@gmail.com}
}

\maketitle
\thispagestyle{empty}

\begin{abstract}
  In this paper, we focus on semi-supervised object detection
  to boost performance of proposal-based object detectors (a.k.a. two-stage object detectors)
  by training on both labeled and unlabeled data.
  However, it is non-trivial to train object detectors on unlabeled data due to the unavailability of ground truth labels.
  To address this problem,
  we present a proposal learning approach
  to learn proposal features and predictions from both labeled and unlabeled data.
  The approach consists of a self-supervised proposal learning module
  and a consistency-based proposal learning module.
  In the self-supervised proposal learning module,
  we present a proposal location loss and a contrastive loss
  to learn context-aware and noise-robust proposal features respectively.
  In the consistency-based proposal learning module,
  we apply consistency losses to both bounding box classification and regression predictions of proposals
  to learn noise-robust proposal features and predictions.
  Our approach enjoys the following benefits:
  1) encouraging more context information to delivered in the proposals learning procedure;
  2) noisy proposal features and enforcing consistency to allow noise-robust object detection;
  3) building a general and high-performance semi-supervised object detection framework, which can be easily adapted to proposal-based object detectors with different backbone architectures.
  Experiments are conducted on the COCO dataset with all available labeled and unlabeled data.
  Results demonstrate that
  our approach consistently improves the performance of fully-supervised baselines.
  In particular, after combining with data distillation \cite{radosavovic2018data},
  our approach improves AP by about 2.0\% and 0.9\% on average compared to fully-supervised baselines and data distillation baselines respectively.
\end{abstract}

\section{Introduction}
\label{sec:intro}

With the giant success of Convolutional Neural Networks (CNNs) \cite{krizhevsky2012imagenet,lecun1998gradient},
great leap forwards have been achieved in object detection \cite{girshick2015fast,girshick2014rich,he2017mask,lin2018focal,liu2016ssd,redmon2016you,ren2017faster}.
However, training accurate object detectors relies on the availability of large scale labeled datasets \cite{everingham2015pascal,lin2014microsoft,russakovsky2015imagenet,shao2019objects365},
which are very expensive and time-consuming to collect.
In addition, training object detectors only on the labeled datasets may limit their detection performance.
By contrast, considering that acquiring unlabeled data is much easier than collecting labeled data,
it is important to explore approaches for the Semi-Supervised Object Detection (SSOD) problem,
\ie, training object detectors on both labeled and unlabeled data,
to boost performance of current state-of-the-art object detectors. 

In this paper, we focus on SSOD for proposal-based object detectors (a.k.a. two-stage object detectors) \cite{girshick2015fast,girshick2014rich,ren2017faster} due to their high performance.
Proposal-based object detectors detect objects by
1) first generating region proposals that may contain objects
and 2) then generating proposal features and predictions
(\ie, bounding box classification and regression predictions).
Specially,
we aim to improve the second stage
by learning proposal features and predictions
from both labeled and unlabeled data.

\begin{figure*}[!tb]
\centering
\includegraphics[width=0.84\linewidth]{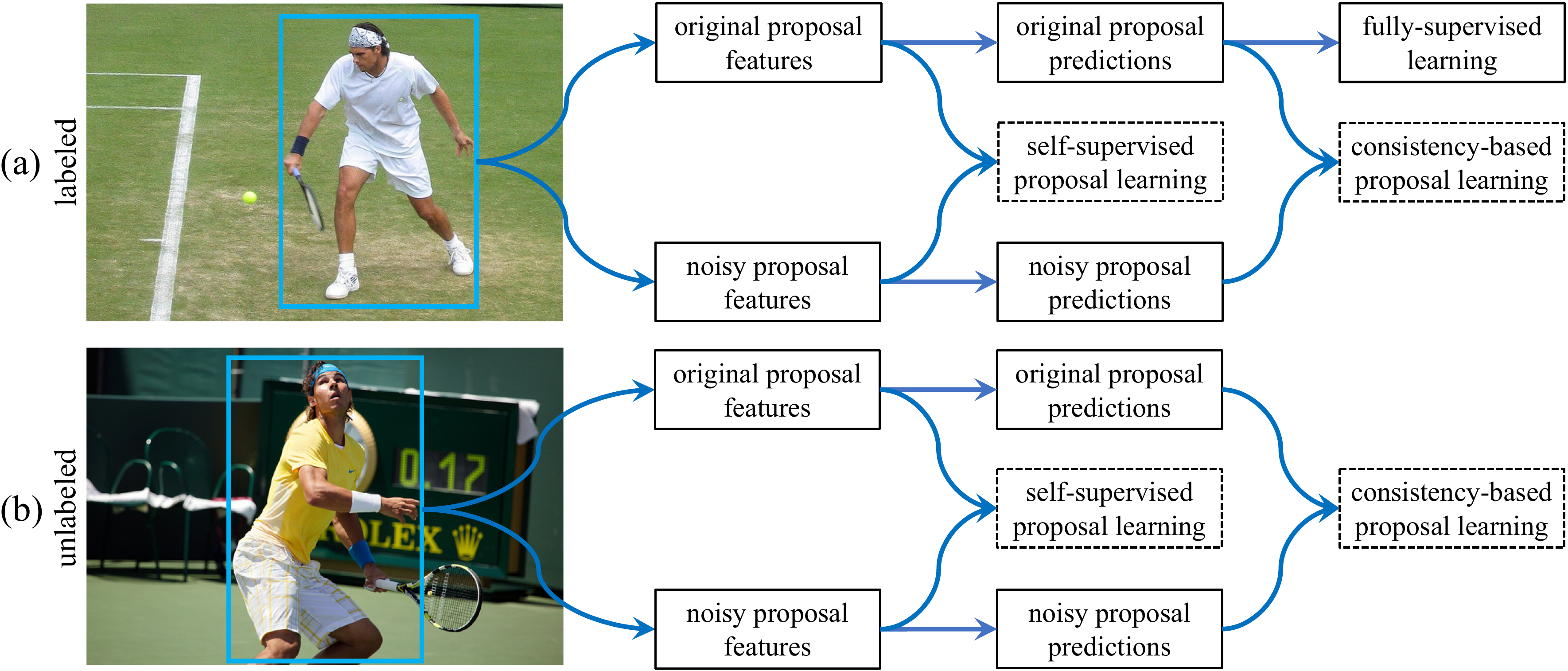}
\caption{The design of our proposal learning approach.
The proposed modules are highlighted in dashed boxes.
Given an image, original/noisy proposal features and predictions are generated.
(a) The standard fully-supervised learning is chosen for labeled data.
(a) and (b) Our proposal learning approach learns proposal features and predictions from both labeled and unlabeled data.
}
\label{fig:overall}
\end{figure*}

For labeled data, it is straightforward to use ground truth labels to get training supervisions.
But for unlabeled data, due to the unavailability of ground truth labels,
we cannot learn proposal features and predictions directly.
To address this problem, apart from the standard fully-supervised learning for labeled data \cite{ren2017faster} shown in Fig.~\ref{fig:overall}~(a),
we present an approach named proposal learning,
which consists of a self-supervised proposal learning module
and a consistency-based proposal learning module,
to learn proposal features and predictions from both labeled and unlabeled data,
see Fig.~\ref{fig:overall}.

Recently, self-supervised learning has shown its efficacy
to learn features from unlabeled data
by defining some pretext tasks \cite{doersch2015unsupervised,he2019momentum,jing2019self,wu2018unsupervised,ye2019unsupervised}.
Our self-supervised proposal learning module uses the same strategy of defining pretext tasks,
inspired by the facts that context is important for object detection \cite{bell2016inside,divvala2009empirical,hu2018relation,mottaghi2014role}
and object detectors should be noise-robust \cite{michaelis2019benchmarking,wang2017fast}.
More precisely,
a proposal location loss and a contrastive loss are presented
to learn context-aware and noise-robust proposal features respectively.
Specifically, the proposal location loss uses proposal location prediction as a pretext task to supervise training,
where a small neural network are attached after proposal features for proposal location prediction.
This loss helps learn context-aware proposal features,
because proposal location prediction requires proposal features understanding some global image information.
At the same time,
the contrastive loss learns noise-robust proposal features
by a simple instance discrimination task \cite{he2019momentum,wu2018unsupervised,ye2019unsupervised},
which ensures that noisy proposals features are closer to their original proposal features than to other proposal features.
In particular, instead of adding noise to images to compute contrastive loss \cite{he2019momentum,ye2019unsupervised},
we add noise to proposal features,
which shares convolutional feature computations for the entire image between noisy proposal feature computations and the original proposal feature computations for training efficiency \cite{girshick2015fast}.

To further train noise-robust object detectors,
our consistency-based proposal learning module
uses consistency losses to ensure that
predictions from noisy proposal features and their original proposal features are consistent.
More precisely, similar to consistency losses for semi-supervised image classification \cite{miyato2018virtual,sajjadi2016regularization,xie2019unsupervised},
a consistency loss for bounding box classification predictions enforces
class predictions from noisy proposal features and their original proposal features to be consistent.
In addition,
a consistency loss for bounding box regression predictions enforces
object location predictions from noisy proposal features and their original proposal features also to be consistent.
With these two consistency losses,
proposal features and predictions are robust to noise.

We apply our approach to Faster R-CNN \cite{ren2017faster} with feature pyramid networks \cite{lin2017feature} and RoIAlign \cite{he2017mask}, using different CNN backbones,
where our proposal learning modules are applied to both labeled and unlabeled data,
as shown in Fig.~\ref{fig:overall}.
We conduct elaborate experiments on the challenging COCO dataset \cite{lin2014microsoft} with all available labeled and unlabeled data,
showing that our approach outperforms fully-supervised baselines consistently.
In particular, when combining with data distillation \cite{radosavovic2018data},
our approach obtains about 2.0\% and 0.9\% absolute AP improvements on average
compared to fully-supervised baselines and data distillation based baselines respectively. 

In summary, we list our main contributions as follows.
\begin{itemize}
\item We present a proposal learning approach
to learn proposal features and predictions from both labeled and unlabeled data.
The approach consists of
1) a self-supervised proposal learning module
which learns context-aware and noise-robust proposal features
by a proposal location loss and a contrastive loss respectively,
and 2) a consistency-based proposal learning module
which learns noise-robust proposal features and predictions
by consistency losses for bounding box classification and regression predictions.
\item On the COCO dataset, our approach surpasses various Faster R-CNN based fully-supervised baselines and data distillation \cite{radosavovic2018data} by about 2.0\% and 0.9\% respectively.
\end{itemize}

\section{Related Work}
\label{sec:related_work}

\begin{figure*}[!tb]
\centering
\includegraphics[width=\linewidth]{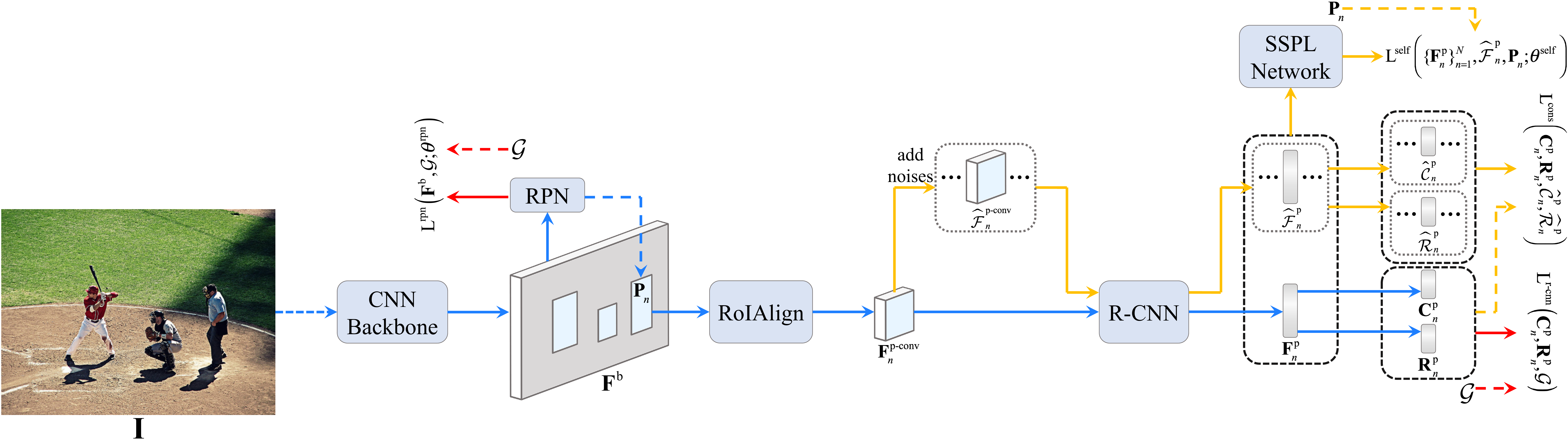}
\caption{The overall framework of our proposal learning approach.
All arrows have forward computations during training,
only the solid ones have back-propagation computations,
the red ones are only utilized for labeled data,
and only the blue ones are utilized during inference.
``RPN'': Region Proposal Network;
``R-CNN'': Region-based CNN;
``SSPL'': Self-Supervised Proposal Learning;
``$\mathbf{I}$'': input image;
``$\mathbf{F}^{\text{b}}$'': image convolutional feature maps;
``$\bm{\theta}^{\text{rpn}}$'': parameters of the RPN;
``$\mathbf{P}_{n}$'': a proposal with its location;
``$\mathbf{F}^{\text{p-conv}}_{n}$'' and ``$\hat{\mathcal{F}}^{\text{p-conv}}_{n}$'':
the original and noisy convolutional feature maps of $\mathbf{P}_{n}$;
``$\mathbf{F}^{\text{p}}_{n}$'' and ``$\hat{\mathcal{F}}^{\text{p}}_{n}$'':
the original and noisy features of $\mathbf{P}_{n}$; 
``$\mathbf{C}^{\text{p}}_{n}, \mathbf{R}^{\text{p}}_{n}$'' and ``$\hat{\mathcal{C}}^{\text{p}}_{n}, \hat{\mathcal{R}}^{\text{p}}_{n}$'':
the original and noisy predictions (bounding box classification and regression predictions) of $\mathbf{P}_{n}$;
``$\bm{\theta}^{\text{self}}$'': parameters of the SSPL network;
``$\mathcal{G}$'': ground truth labels;
``$\mathrm{L}^{\text{rpn}}\left(\mathbf{F}^{\text{b}}, \mathcal{G}; \bm{\theta}^{\text{rpn}}\right)$'': RPN loss;
``$\mathrm{L}^{\text{r-cnn}}\left(\mathbf{C}^{\text{p}}_{n}, \mathbf{R}^{\text{p}}_{n}, \mathcal{G}\right)$'': R-CNN loss;
``$\mathrm{L}^{\text{self}}\left(\{\mathbf{F}^{\text{p}}_{n}\}_{n=1}^{N}, \hat{\mathcal{F}}^{\text{p}}_{n}, \mathbf{P}_{n}; \bm{\theta}^{\text{self}}\right)$'':
SSPL loss;
``$\mathrm{L}^{\text{cons}}\left(\mathbf{C}^{\text{p}}_{n}, \mathbf{R}^{\text{p}}_{n}, \hat{\mathcal{C}}^{\text{p}}_{n}, \hat{\mathcal{R}}^{\text{p}}_{n}\right)$'':
consistency-based proposal learning loss.
See Section~\ref{sec:framework} for more details.
Best viewed in color.
}
\label{fig:framework}
\end{figure*}

\noindent\textbf{Object detection} is one of the most important tasks in computer vision
and has received considerable attention in recent years \cite{cai2018cascade,girshick2015fast,girshick2014rich,he2017mask,hu2018relation,huang2019mask,lin2017feature,lin2018focal,liu2016ssd,redmon2016you,ren2017faster} \cite{tang2019object,wang2019deep,wang2017fast,zhang2018single}.
One popular direction for recent object detection is proposal-based object detectors (a.k.a. two-stage object detectors) \cite{girshick2015fast,girshick2014rich,he2017mask,lin2017feature,ren2017faster},
which perform object detection
by first generating region proposals
and then generating proposal features and predictions.
Very promising results are obtained by these proposal-based approaches.
In this work, we are also along the line of proposal-based object detectors.
But unlike previous approaches training object detectors only on labeled data,
we train object detectors on both labeled data and unlabeled data,
and present a proposal learning approach to achieve our goal.
In addition, Wang \etal\ \cite{wang2017fast} also add noise to proposal features to train noise-robust object detectors.
They focus on generating hard noisy proposal features and present adversarial networks based approaches,
and still train object detectors only on labeled data.
Unlike their approach,
we add noise to proposal features
to learn noise-robust proposal features and predictions
from both labeled and unlabeled data
by our proposal learning approach.

\vspace{0.1cm}
\noindent\textbf{Self-supervised learning}
learns features from unlabeled data by some defined pretext tasks \cite{doersch2015unsupervised,gidaris2018unsupervised,he2019momentum,kolesnikov2019revisiting,pathak2016context,ye2019unsupervised,zhang2016colorful}.
For example, Doersch \etal\ \cite{doersch2015unsupervised} predicts the position of one patch relative to another patch in the same image.
Gidaris \etal\ \cite{gidaris2018unsupervised} randomly rotate images and predict the rotation of images.
Some recent works \cite{he2019momentum,wu2018unsupervised,ye2019unsupervised} use an instance discrimination task to match features from noisy images with features from their original images.
Please see the recent survey \cite{jing2019self} for more self-supervised learning approaches.
Our self-supervised proposal learning module applies self-supervised learning approaches to learn proposal features from both labeled and unlabeled data.
Inspired by unsupervised feature learning task \cite{doersch2015unsupervised} and the instance discrimination task \cite{he2019momentum,wu2018unsupervised,ye2019unsupervised} of image level, we introduce self-supervised learning to SSOD
by designing a proposal location loss and contrastive loss on object proposals.

\vspace{0.1cm}
\noindent\textbf{Semi-supervised learning}
trains models on both labeled and unlabeled data.
There are multiple strategies for semi-supervised learning,
such as self-training \cite{yarowsky1995unsupervised}, co-training \cite{blum1998combining,qiao2018deep,zhou2019semi}, label propagation \cite{zhu2002learning}, \etc\
Please see \cite{chapelle2010semi} for an extensive review.
Recently, many works use consistency losses for semi-supervised image classification \cite{athiwaratkun2018there,laine2016temporal,miyato2018virtual,sajjadi2016regularization,tarvainen2017mean,xie2019unsupervised},
by enforcing class predictions from noisy inputs and their original inputs to be consistent,
where noise is added to input images or intermediate features.
Here we add noise to proposal features for efficiency
and apply consistency losses to both class and object location predictions of proposals.
Zhai \etal\ \cite{zhai2019s4l} also suggest to benefit semi-supervised image classification from self-supervised learning.
Here we further apply self-supervised learning to SSOD
by a self-supervised proposal learning module.

\vspace{0.1cm}
\noindent\textbf{Semi-supervised object detection}
applies semi-supervised learning to object detection.
There are some SSOD works with different settings \cite{cinbis2016weakly,gao2019note,hoffman2014lsda,tang2016large}.
For example, Cinbis \etal\ \cite{cinbis2016weakly} train object detectors on data with either bounding box labels or image-level class labels.
Hoffman \etal\ \cite{hoffman2014lsda} and Tang \etal\ \cite{tang2016large} train object detectors on data
with bounding box labels for some classes
and image-level class labels for other classes.
Gao \etal\ \cite{gao2019note} train object detectors on data
with bounding box labels for some classes
and either image-level labels or bounding box labels for other classes.
Unlike their settings,
in this work we explore the more general semi-supervised setting,
\ie, training object detectors on data
which either have bounding box labels or are totally unlabeled,
similar to the standard semi-supervised learning setting \cite{chapelle2010semi}.
Jeong \etal\ \cite{jeong2019consistency} also use consistency losses for SSOD by adding noise to images.
Unlike their approach, we add noise to proposal features instead of images
and present a self-supervised proposal learning module.
In addition, all these works mainly conduct experiments on 
simulated labeled/unlabeled data by splitting a fully annotated dataset
and thus cannot fully utilize the available labeled data \cite{radosavovic2018data}.
Our work follows the setting in \cite{radosavovic2018data}
which trains object detectors on both labeled and unlabeled data
and uses all labeled COCO data during training.
Unlike the data distillation approach presented in \cite{radosavovic2018data} which can be viewed as a self-training-based approach,
our approach combines self-supervised learning and consistency losses for SSOD.
In experiments, we will show that our approach and the data distillation approach are complementary to some extent.

\section{Approach}
\label{sec:approach}

In this section,
we will first give the definition of our Semi-Supervised Object Detection (SSOD) problem (see Section~\ref{sec:problem}),
then describe our overall framework (see Section~\ref{sec:framework}),
and finally introduce our proposal learning approach,
consisting of the self-supervised proposal learning module (see Section~\ref{sec:sspl})
and consistency-based proposal learning module (see Section~\ref{sec:cbpl}).
If not specified,the contents we described here are the training procedures
since we aim to train object detectors under the SSOD setting.

\subsection{Problem Definition}
\label{sec:problem}

In SSOD,
a set of labeled data $\mathcal{D}^{\text{l}} = \{(\mathbf{I}, \mathcal{G})\}$ and a set of unlabeled data $\mathcal{D}^{\text{u}} = \{\mathbf{I}\}$ are given,
where $\mathbf{I}$ and $\mathcal{G}$ denote
an image and ground truth labels respectively.
In object detection, $\mathcal{G}$ consists of
a set of objects with locations and object classes.
Our goal of SSOD is to train object detectors on both labeled data $\mathcal{D}^{\text{l}}$ and unlabeled data $\mathcal{D}^{\text{u}}$.

\subsection{The Overall Framework}
\label{sec:framework}

The overall framework of our approach is shown in Fig.~\ref{fig:framework}.
As in standard proposal-based object detectors \cite{lin2017feature,ren2017faster},
during the forward process,
first, an input image $\mathbf{I}$ is fed into a CNN backbone (\eg, ResNet-50 \cite{he2016deep} with feature pyramid networks \cite{lin2017feature}) with parameters $\bm{\theta}^{\text{b}}$,
which produces image convolutional feature maps $\mathbf{F}^{\text{b}}(\mathbf{I}; \bm{\theta}^{\text{b}})$.
Then, a Region Proposal Network (RPN) with parameters $\bm{\theta}^{\text{rpn}}$ takes $\mathbf{F}^{\text{b}}(\mathbf{I}; \bm{\theta}^{\text{b}})$ as inputs to generate region proposals $\mathcal{P}\left(\mathbf{F}^{\text{b}}(\mathbf{I}; \bm{\theta}^{\text{b}}); \bm{\theta}^{\text{rpn}}\right)$.
We use $\mathbf{F}^{\text{b}}, \mathcal{P}$ later for simplification, dropping the dependence on $\mathbf{I}, \bm{\theta}^{\text{b}}, \mathbf{F}^{\text{b}}, \bm{\theta}^{\text{rpn}}$.
Next, an RoIAlign \cite{he2017mask} layer takes each proposal $\mathbf{P}_{n} = (x_{n}, y_{n}, w_{n}, h_{n}) \in \mathcal{P}$ and $\mathbf{F}^{\text{b}}$ as inputs
to extract proposal convolutional feature maps $\mathbf{F}^{\text{p-conv}}_{n}$ (simplification of $\mathbf{F}^{\text{p-conv}}(\mathbf{F}^{\text{b}}, \mathbf{P}_{n})$, dropping the dependence on $\mathbf{F}^{\text{b}}, \mathbf{P}_{n}$),
where $(x_{n}, y_{n}, w_{n}, h_{n})$ denotes the location of the $i^{\text{th}}$ proposal $\mathbf{P}_{n}$,
$i \in \{1, 2, ..., N\}$,
and $N$ is the number of proposals in $\mathcal{P}$.
After that, $\mathbf{F}^{\text{p-conv}}_{n}$ is fed into a Region-based CNN (R-CNN)
to generate proposal features $\mathbf{F}^{\text{p}}(\mathbf{F}^{\text{p-conv}}_{n}; \bm{\theta}^{\text{r-cnn}})$ and predictions
(\ie, bounding box classification predictions $\mathbf{C}^{\text{p}}\left(\mathbf{F}^{\text{p}}(\mathbf{F}^{\text{p-conv}}_{n}; \bm{\theta}^{\text{r-cnn}}); \bm{\theta}^{\text{cls}}\right)$ and bounding box regression predictions $\mathbf{R}^{\text{p}}\left(\mathbf{F}^{\text{p}}(\mathbf{F}^{\text{p-conv}}_{n}; \bm{\theta}^{\text{r-cnn}}); \bm{\theta}^{\text{reg}}\right)$),
where $\bm{\theta}^{\text{r-cnn}}, \bm{\theta}^{\text{cls}}$, and $\bm{\theta}^{\text{reg}}$ denote
parameters of the R-CNN to generate proposal features, bounding box classification predictions, and bounding box regression predictions, respectively.
We use $\mathbf{F}^{\text{p}}_{n}, \mathbf{C}^{\text{p}}_{n}, \mathbf{R}^{\text{p}}_{n}$ later for simplification, dropping the dependence on $\mathbf{F}^{\text{p-conv}}_{n}, \bm{\theta}^{\text{r-cnn}}, \mathbf{F}^{\text{p}}_{n}, \bm{\theta}^{\text{cls}}, \bm{\theta}^{\text{reg}}$.

For each labeled data $(\mathbf{I}, \mathcal{G}) \in \mathcal{D}^{\text{l}}$,
it is straightforward to train object detectors according to the standard fully-supervised learning loss defined in Eq.~(\ref{equ:l_labeled}),
where the first and second terms denote the RPN loss and R-CNN loss respectively.
This loss is optimized \wrt\ $\bm{\theta}^{\text{b}}, \bm{\theta}^{\text{rpn}}, \bm{\theta}^{\text{r-cnn}}, \bm{\theta}^{\text{cls}}, \bm{\theta}^{\text{reg}}$ to train object detectors
during the back-propagation process.
More details of the loss function can be found in \cite{ren2017faster}.
\begin{equation}
\label{equ:l_labeled}
\begin{aligned}
   &\mathrm{L}^{\text{sup}}\left(\mathbf{I}, \mathcal{G}; \bm{\theta}^{\text{b}}, \bm{\theta}^{\text{rpn}}, \bm{\theta}^{\text{r-cnn}}, \bm{\theta}^{\text{cls}}, \bm{\theta}^{\text{reg}}\right)\\
   = \ &\mathrm{L}^{\text{rpn}}\left(\mathbf{I}, \mathcal{G}; \bm{\theta}^{\text{b}}, \bm{\theta}^{\text{rpn}}\right) + \frac{1}{N} \mathop{\sum} \limits_{n}\mathrm{L}^{\text{r-cnn}}\left(\mathbf{I}, \mathbf{P}_{n}, \mathcal{G}, \bm{\theta}^{\text{b}}, \bm{\theta}^{\text{r-cnn}}, \bm{\theta}^{\text{cls}}, \bm{\theta}^{\text{reg}}\right) \\
   = \ &\mathrm{L}^{\text{rpn}}\left(\mathbf{F}^{\text{b}}, \mathcal{G}; \bm{\theta}^{\text{rpn}}\right)
   + \frac{1}{N} \mathop{\sum} \limits_{n}\mathrm{L}^{\text{r-cnn}}\left(\mathbf{C}^{\text{p}}_{n}, \mathbf{R}^{\text{p}}_{n}, \mathcal{G}\right).
\end{aligned}
\end{equation}

However, for unlabeled data $\mathbf{I} \in \mathcal{D}^{\text{u}}$,
there is no available ground truth labels $\mathcal{G}$.
Thus we cannot use Eq.~(\ref{equ:l_labeled}) to train object detectors on $\mathcal{D}^{\text{u}}$.
To train object detectors also on $\mathcal{D}^{\text{u}}$,
we present a proposal learning approach,
consisting of a self-supervised proposal learning module
and a consistency-based proposal learning module,
to also learn proposal features (\ie, $\mathbf{F}^{\text{p}}_{n}$) and predictions (\ie, $\mathbf{C}^{\text{p}}_{n}, \mathbf{R}^{\text{p}}_{n}$) from $\mathcal{D}^{\text{u}}$.
It is possible to also benefit RPN from $\mathcal{D}^{\text{u}}$.
We only focus on the R-CNN-related parts,
because 1) the final object detection results are from R-CNN-related parts
and thus improving the R-CNN-related parts will benefit object detectors directly;
2) gradients will also be back-propagated from the R-CNN-related parts to the CNN backbone to learn better $\mathbf{F}^{b}$,
which could potentially improve RPN.

For the $n^{\text{th}}$ proposal $\mathbf{P}_{n}$ of image $\mathbf{I}$,
during the forward process,
we first generate a set of noisy proposal features $\hat{\mathcal{F}}^{\text{p}}_{n} = \{\hat{\mathbf{F}}^{\text{p}}_{nk}\}_{k=1}^{K}$
and predictions $\hat{\mathcal{C}}^{\text{p}}_{n} = \{\hat{\mathbf{C}}^{\text{p}}_{nk}\}_{k=1}^{K}, \hat{\mathcal{R}}^{\text{p}}_{n} = \{\hat{\mathbf{R}}^{\text{p}}_{nk}\}_{k=1}^{K}$,
where $K$ denotes the number of noisy proposal features for each $\mathbf{P}_{n}$.
As we stated in Section~\ref{sec:intro},
we add noise to proposal features to share convolutional feature computations for the CNN backbone (\ie, $\mathbf{F}^{\text{b}}$)
between noisy proposal feature computations and the original proposal feature computations for efficiency.
More specifically,
we add random noise $\{\epsilon_{nk}\}_{k=1}^{K}$ to proposal convolutional feature maps $\mathbf{F}^{\text{p-conv}}_{n}$,
which generates a set of noisy proposal convolutional feature maps $\hat{\mathcal{F}}^{\text{p-conv}}_{n} = \{\hat{\mathbf{F}}^{\text{p-conv}}(\mathbf{F}^{\text{p-conv}}_{n}, \epsilon_{nk})\}_{k=1}^{K}$,
see Fig.~\ref{fig:framework}.
We use $\hat{\mathbf{F}}^{\text{p-conv}}_{nk}$ for simplification, dropping the dependence on $\mathbf{F}^{\text{p-conv}}_{n}, \epsilon_{nk}$.
Similar to the procedure to generate $\mathbf{F}^{\text{p}}_{n}, \mathbf{C}^{\text{p}}_{n}, \mathbf{R}^{\text{p}}_{n}$,
the noisy proposal feature maps are fed into the R-CNN to generate the noisy proposal features $\hat{\mathcal{F}}^{\text{p}}_{n} = \{\hat{\mathbf{F}}^{\text{p}}_{nk}\}_{k=1}^{K}$
and predictions $\hat{\mathcal{C}}^{\text{p}}_{n} = \{\hat{\mathbf{C}}^{\text{p}}_{nk}\}_{k=1}^{K}, \hat{\mathcal{R}}^{\text{p}}_{n} = \{\hat{\mathbf{R}}^{\text{p}}_{nk}\}_{k=1}^{K}$
(we drop the dependence on $\hat{\mathbf{F}}^{\text{p-conv}}_{nk}, \bm{\theta}^{\text{r-cnn}}, \hat{\mathbf{F}}^{\text{p}}_{nk}, \bm{\theta}^{\text{cls}}, \bm{\theta}^{\text{reg}}$ for notation simplification).

For Self-Supervised Proposal Learning (SSPL),
as shown in Fig.~\ref{fig:framework},
during the forward process,
we pass the original proposal features $\mathbf{F}^{\text{p}}_{n}$
and the noisy proposal features $\hat{\mathcal{F}}^{\text{p}}_{n}$
through a small SSPL network with parameters $\bm{\theta}^{\text{self}}$.
The outputs of the SSPL network and the proposal location $\mathbf{P}_{n} = (x_{n}, y_{n}, w_{n}, h_{n})$
are taken as inputs to compute SSPL loss
$\mathrm{L}^{\text{self}}\left(\{\mathbf{F}^{\text{p}}_{n}\}_{n=1}^{N}, \hat{\mathcal{F}}^{\text{p}}_{n}, \mathbf{P}_{n}; \bm{\theta}^{\text{self}}\right)$
which is defined in later Eq.~(\ref{equ:l_sspl}).
Since this loss does not take any ground truth labels $\mathcal{G}$ as inputs,
by optimizing this loss \wrt\ $\mathbf{F}^{\text{p}}_{n}, \hat{\mathcal{F}}^{\text{p}}_{n}, \bm{\theta}^{\text{self}}$,
\ie, $\bm{\theta}^{\text{b}}, \bm{\theta}^{\text{r-cnn}}, \bm{\theta}^{\text{self}}$,
during the back-propagation process,
we can learn proposal features also from unlabeled data.
We will give more details about this module in Section~\ref{sec:sspl}.

For consistency-based proposal learning,
as shown in Fig.~\ref{fig:framework},
during the forward process
the original proposal predictions $\mathbf{C}^{\text{p}}_{n}, \mathbf{R}^{\text{p}}_{n}$
and the noisy proposal predictions $\hat{\mathcal{C}}^{\text{p}}_{n}, \hat{\mathcal{R}}^{\text{p}}_{n}$
are taken as inputs to compute loss
$\mathrm{L}^{\text{cons}}\left(\mathbf{C}^{\text{p}}_{n}, \mathbf{R}^{\text{p}}_{n}, \hat{\mathcal{C}}^{\text{p}}_{n}, \hat{\mathcal{R}}^{\text{p}}_{n}\right)$
which is defined in later Eq.~(\ref{equ:l_cbpl}).
Following \cite{miyato2018virtual,xie2019unsupervised},
this loss is optimized \wrt\ $\hat{\mathcal{C}}^{\text{p}}_{n}, \hat{\mathcal{R}}^{\text{p}}_{n}$ (not \wrt\ $\mathbf{C}^{\text{p}}_{n}, \mathbf{R}^{\text{p}}_{n}$),
\ie, $\bm{\theta}^{\text{b}}, \bm{\theta}^{\text{r-cnn}}, \bm{\theta}^{\text{cls}}, \bm{\theta}^{\text{reg}}$,
during the back-propagation process.
Computing this loss does not require any ground truth labels $\mathcal{G}$,
and thus we can learn proposal features and predictions also from unlabeled data.
We will give more details about this module in Section~\ref{sec:cbpl}.

We apply the standard fully-supervised loss defined in Eq.~(\ref{equ:l_labeled}) to labeled data $\mathcal{D}^{\text{l}}$,
and the self-supervised proposal learning loss $\mathrm{L}^{\text{self}}\left(\{\mathbf{F}^{\text{p}}_{n}\}_{n=1}^{N}, \hat{\mathcal{F}}^{\text{p}}_{n}, \mathbf{P}_{n}; \bm{\theta}^{\text{self}}\right)$
and the consistency-based proposal learning loss $\mathrm{L}^{\text{cons}}\left(\mathbf{C}^{\text{p}}_{n}, \mathbf{R}^{\text{p}}_{n}, \hat{\mathcal{C}}^{\text{p}}_{n}, \hat{\mathcal{R}}^{\text{p}}_{n}\right)$
to unlabeled data $\mathcal{D}^{\text{u}}$.
The object detectors are trained on $\mathcal{D}^{\text{l}}, \mathcal{D}^{\text{u}}$
by optimizing the loss Eq.~(\ref{equ:l_all})
\wrt\ $\bm{\theta}^{\text{b}}, \bm{\theta}^{\text{rpn}}, \bm{\theta}^{\text{r-cnn}}, \bm{\theta}^{\text{cls}}, \bm{\theta}^{\text{reg}}, \bm{\theta}^{\text{self}}$
during the back-propagation process.
\begin{equation}
\label{equ:l_all}
\begin{aligned}
  &\mathrm{L}\left(\mathbf{I}, \mathcal{G}; \bm{\theta}^{\text{b}}, \bm{\theta}^{\text{rpn}}, \bm{\theta}^{\text{r-cnn}}, \bm{\theta}^{\text{cls}}, \bm{\theta}^{\text{reg}}, \bm{\theta}^{\text{self}}\right)\\
  = &\frac{1}{|\mathcal{D}^{\text{l}}|} \mathop{\sum} \limits_{(\mathbf{I}, \mathcal{G}) \in \mathcal{D}^{\text{l}}} \mathrm{L}^{\text{sup}}\left(\mathbf{I}, \mathcal{G}; \bm{\theta}^{\text{b}}, \bm{\theta}^{\text{rpn}}, \bm{\theta}^{\text{r-cnn}}, \bm{\theta}^{\text{cls}}, \bm{\theta}^{\text{reg}}\right)\\
  &+ \frac{1}{|\mathcal{D}^{\text{u}}|} \mathop{\sum}_{\mathbf{I} \in \mathcal{D}^{\text{u}}} \frac{1}{N} \mathop{\sum}_{n}
  \mathrm{L}^{\text{self}}\left(\{\mathbf{F}^{\text{p}}_{n}\}_{n=1}^{N}, \hat{\mathcal{F}}^{\text{p}}_{n}, \mathbf{P}_{n}; \bm{\theta}^{\text{self}}\right)\\
  &+ \frac{1}{|\mathcal{D}^{\text{u}}|} \mathop{\sum}_{\mathbf{I} \in \mathcal{D}^{\text{u}}} \frac{1}{N} \mathop{\sum}_{n}
  \mathrm{L}^{\text{cons}}\left(\mathbf{C}^{\text{p}}_{n}, \mathbf{R}^{\text{p}}_{n}, \hat{\mathcal{C}}^{\text{p}}_{n}, \hat{\mathcal{R}}^{\text{p}}_{n}\right).
\end{aligned}
\end{equation}
We can also apply the self-supervised and consistency-based proposal learning losses to both labeled and unlabeled data,
following the semi-supervised learning works \cite{laine2016temporal,miyato2018virtual,zhai2019s4l}.
Then the overall loss is written as Eq.~(\ref{equ:l_all_new}).
\begin{equation}
\label{equ:l_all_new}
\begin{aligned}
   &\mathrm{L}\left(\mathbf{I}, \mathcal{G}; \bm{\theta}^{\text{b}}, \bm{\theta}^{\text{rpn}}, \bm{\theta}^{\text{r-cnn}}, \bm{\theta}^{\text{cls}}, \bm{\theta}^{\text{reg}}, \bm{\theta}^{\text{self}}\right)\\
  = &\frac{1}{|\mathcal{D}^{\text{l}}|} \mathop{\sum} \limits_{(\mathbf{I}, \mathcal{G}) \in \mathcal{D}^{\text{l}}} \mathrm{L}^{\text{sup}}\left(\mathbf{I}, \mathcal{G}; \bm{\theta}^{\text{b}}, \bm{\theta}^{\text{rpn}}, \bm{\theta}^{\text{r-cnn}}, \bm{\theta}^{\text{cls}}, \bm{\theta}^{\text{reg}}\right)\\
  &+ \frac{1}{|\mathcal{D}^{\text{l}}| + |\mathcal{D}^{\text{u}}|} \mathop{\sum}_{\mathbf{I} \in \mathcal{D}^{\text{l}}, \mathcal{D}^{\text{u}}} \frac{1}{N} \mathop{\sum}_{n}
  \mathrm{L}^{\text{self}}\left(\{\mathbf{F}^{\text{p}}_{n}\}_{n=1}^{N}, \hat{\mathcal{F}}^{\text{p}}_{n}, \mathbf{P}_{n}; \bm{\theta}^{\text{self}}\right)\\
  &+ \frac{1}{|\mathcal{D}^{\text{l}}| + |\mathcal{D}^{\text{u}}|} \mathop{\sum}_{\mathbf{I} \in \mathcal{D}^{\text{l}}, \mathcal{D}^{\text{u}}} \frac{1}{N} \mathop{\sum}_{n}
  \mathrm{L}^{\text{cons}}\left(\mathbf{C}^{\text{p}}_{n}, \mathbf{R}^{\text{p}}_{n}, \hat{\mathcal{C}}^{\text{p}}_{n}, \hat{\mathcal{R}}^{\text{p}}_{n}\right).
\end{aligned}
\end{equation}

During inference,
we simply keep the parts of the standard proposal-based object detectors, see the blue arrows in Fig.~\ref{fig:framework}.
Therefore, our approach does not bring any extra inference computations.

\subsection{Self-Supervised Proposal Learning}
\label{sec:sspl}

Previous works have shown that object detectors can benefit from context \cite{bell2016inside,divvala2009empirical,hu2018relation,mottaghi2014role}
and should be noise-robust \cite{michaelis2019benchmarking,wang2017fast}.
Our self-supervised proposal learning module
uses a proposal location loss and a contrastive loss
to learn context-aware and noise-robust proposal features respectively.

To compute proposal location loss,
we use proposal location prediction as the pretext task,
inspired by the approach in \cite{doersch2015unsupervised}.
More specifically,
we pass $\mathbf{F}^{\text{p}}_{n}, \hat{\mathcal{F}}^{\text{p}}_{n}$ through two fully-connected layers with parameters $\bm{\theta}^{\text{self-loc}}$ and a sigmoid layer to compute location predictions $\mathbf{L}^{\text{p}}_{n}, \hat{\mathcal{L}}^{\text{p}}_{n} = \{\hat{\mathbf{L}}^{\text{p}}_{nk}\}_{k=1}^{K}$,
where the numbers of the outputs of the two fully-connected layers are $1024$ and $4$ respectively.
Here we drop the dependence on $\mathbf{F}^{\text{p}}_{n}, \hat{\mathcal{F}}^{\text{p}}_{n}, \bm{\theta}^{\text{self-loc}}$ for notation simplification.
Then we use $\ell_{2}$ distance to compute proposal location loss,
see Eq.~(\ref{equ:l_location}),
where $\tilde{\mathbf{P}}_{n} = (x_{n} / W, y_{n} / H, w_{n} / W, h_{n} / H)$ is a normalized version of $\mathbf{P}_{n}$,
and $W, H$ denote the width and height of image $\mathbf{I}$ respectively.
\begin{equation}
\label{equ:l_location}
\begin{aligned}
   &\mathrm{L}^{\text{self-loc}}\left(\mathbf{F}^{\text{p}}_{n}, \hat{\mathcal{F}}^{\text{p}}_{n}, \mathbf{P}_{n}; \bm{\theta}^{\text{self-loc}}\right)\\
   = \ &\mathrm{L}^{\text{self-loc}}\left(\mathbf{L}^{\text{p}}_{n}, \hat{\mathcal{L}}^{\text{p}}_{n}, \mathbf{P}_{n}\right)\\
   = \ &\frac{1}{K + 1}\left(\|\mathbf{L}^{\text{p}}_{n} - \tilde{\mathbf{P}}_{n}\|_{2}^{2} + \mathop{\sum} \limits_{k}\|\hat{\mathbf{L}}^{\text{p}}_{nk} - \tilde{\mathbf{P}}_{n}\|_{2}^{2}\right).
\end{aligned}
\end{equation}
By optimizing this loss \wrt\ $\mathbf{F}^{\text{p}}_{n}, \hat{\mathcal{F}}^{\text{p}}_{n}, \bm{\theta}^{\text{self-loc}}$,
\ie, $\bm{\theta}^{\text{b}}, \bm{\theta}^{\text{r-cnn}}, \bm{\theta}^{\text{self-loc}}$,
we can learn context-aware proposal features
because predicting proposal location in an image
requires proposal features understanding some global information of the image.
We do not use the relative patch location prediction task \cite{doersch2015unsupervised} directly,
because images are large and there are always multiple objects in the same image for object detection,
which makes relative patch location prediction hard to be solved.

To compute contrastive loss,
we use instance discrimination as the pretext task, following \cite{he2019momentum,wu2018unsupervised,ye2019unsupervised}.
More specifically,
we first use a fully-connected layer with parameters $\bm{\theta}^{\text{self-cont}}$ and an $\ell_{2}$ normalization layer to project $\mathbf{F}^{\text{p}}_{n}, \hat{\mathcal{F}}^{\text{p}}_{n}$
to embedded proposal features $\mathbf{F}^{\text{embed}}_{n}, \hat{\mathcal{F}}^{\text{embed}}_{n} = \{\hat{\mathbf{F}}^{\text{embed}}_{nk}\}_{k=1}^{K}$
(dropping the dependence on $\mathbf{F}^{\text{p}}_{n}, \hat{\mathcal{F}}^{\text{p}}_{n}, \bm{\theta}^{\text{self-cont}}$),
where the numbers of the outputs of the fully-connected layer is $128$.
Then the contrastive loss is written as Eq.~(\ref{equ:l_contrastive}),
where $\tau$ is a temperature hyper-parameter.
\begin{equation}
\label{equ:l_contrastive}
\begin{aligned}
   &\mathrm{L}^{\text{self-cont}}\left(\{\mathbf{F}^{\text{p}}_{n}\}_{n=1}^{N}, \hat{\mathcal{F}}^{\text{p}}_{n}; \bm{\theta}^{\text{self-cont}}\right)\\
   = \ &\mathrm{L}^{\text{self-cont}}\left(\{\mathbf{F}^{\text{embed}}_{n}\}_{n=1}^{N}, \hat{\mathcal{F}}^{\text{embed}}_{n}\right)\\
   = \ &- \frac{1}{K} \mathop{\sum} \limits_{k} \log\frac{\exp({(\hat{\mathbf{F}}^{\text{embed}}_{nk}})^{\text{T}} \mathbf{F}^{\text{embed}}_{n} / \tau)}{\sum_{n^{\prime}}\exp({(\hat{\mathbf{F}}^{\text{embed}}_{nk}})^{\text{T}} \mathbf{F}^{\text{embed}}_{n^{\prime}} / \tau)}.
\end{aligned}
\end{equation}
By optimizing this loss \wrt\ $\mathbf{F}^{\text{p}}_{n}, \hat{\mathcal{F}}^{\text{p}}_{n}, \bm{\theta}^{\text{self-cont}}$,
\ie, $\bm{\theta}^{\text{b}}, \bm{\theta}^{\text{r-cnn}}, \bm{\theta}^{\text{self-cont}}$,
noisy proposal features are enforced to be closer to their original proposal features
than to other proposal features,
which learns noise-robust proposal features and thus learns noise-robust object detectors.

By combining the proposal location loss in Eq.~(\ref{equ:l_location}) and the contrastive loss in Eq.~(\ref{equ:l_contrastive}),
the overall self-supervised proposal learning loss is written as Eq.~(\ref{equ:l_sspl}),
where $\lambda^{\text{self-loc}}, \lambda^{\text{self-cont}}$ are loss weights and $\bm{\theta}^{\text{self}} = \{\bm{\theta}^{\text{self-loc}}, \bm{\theta}^{\text{self-cont}}\}$.
This loss is optimized \wrt\ $\bm{\theta}^{\text{b}}, \bm{\theta}^{\text{r-cnn}}, \bm{\theta}^{\text{self}}$ to learn proposal features.
\begin{equation}
\label{equ:l_sspl}
\begin{aligned}
   &\mathrm{L}^{\text{self}}\left(\{\mathbf{F}^{\text{p}}_{n}\}_{n=1}^{N}, \hat{\mathcal{F}}^{\text{p}}_{n}; \bm{\theta}^{\text{self}}\right)\\
   = \ & \lambda^{\text{self-loc}} \mathrm{L}^{\text{self-loc}}\left(\mathbf{F}^{\text{p}}_{n}, \hat{\mathcal{F}}^{\text{p}}_{n}, \mathbf{P}_{n}; \bm{\theta}^{\text{self-loc}}\right) \\
   & + \lambda^{\text{self-cont}} \mathrm{L}^{\text{self-cont}}\left(\{\mathbf{F}^{\text{p}}_{n}\}_{n=1}^{N}, \hat{\mathcal{F}}^{\text{p}}_{n}; \bm{\theta}^{\text{self-cont}}\right).
\end{aligned}
\end{equation}

\subsection{Consistency-Based Proposal Learning}
\label{sec:cbpl}

To further train noise-robust object detectors,
we apply consistency losses \cite{miyato2018virtual,sajjadi2016regularization,xie2019unsupervised} to ensure consistency between noisy proposal predictions and their original proposal predictions.
More precisely,
we apply consistency losses to both bounding box classification and regression predictions.

For consistency loss for bounding box classification predictions $\mathbf{C}^{\text{p}}_{n}, \hat{\mathcal{C}}^{\text{p}}_{n}$,
we use KL divergence as the loss to enforce class predictions from noisy proposals and their original proposals to be consistent,
follwing \cite{miyato2018virtual,xie2019unsupervised}, see Eq.~(\ref{equ:l_cls}).
\begin{equation}
\label{equ:l_cls}
\begin{aligned}
   \mathrm{L}^{\text{cons-cls}}\left(\mathbf{C}^{\text{p}}_{n}, \hat{\mathcal{C}}^{\text{p}}_{n}\right)
   = \frac{1}{K} \mathop{\sum} \limits_{k} \operatorname{KL}\left(\mathbf{C}^{\text{p}}_{n}\|\hat{\mathbf{C}}^{\text{p}}_{nk}\right).
\end{aligned}
\end{equation}

Unlike image classification containing only classification results,
object detection also predicts object locations.
To further ensure proposal prediction consistency,
we compute consistency loss in Eq.~(\ref{equ:l_reg})
to enforce object location predictions from noisy proposals and their original proposals to be consistent.
Here we use the standard bounding box regression loss, \ie, smoothed $\ell_{1}$ loss \cite{girshick2015fast}.
We only selected the easiest noisy proposal feature to compute this loss for training stability.
\begin{equation}
\label{equ:l_reg}
\begin{aligned}
   \mathrm{L}^{\text{cons-reg}}\left(\mathbf{R}^{\text{p}}_{n}, \hat{\mathcal{R}}^{\text{p}}_{n}\right)
   = \mathop{\min} \limits_{k} \left( \operatorname{smooth_{\ell_{1}}}\left(\mathbf{R}^{\text{p}}_{n} - \hat{\mathbf{R}}^{\text{p}}_{nk}\right)\right).
\end{aligned}
\end{equation}

By combining the consistency losses for bounding box classification predictions in Eq.~(\ref{equ:l_cls})
and for bounding box regression predictions in Eq.~(\ref{equ:l_reg}),
the overall consistency-based proposal learning loss is written as Eq.~(\ref{equ:l_cbpl}),
where $\lambda^{\text{cons-cls}}, \lambda^{\text{cons-reg}}$ are loss weights.
Following \cite{miyato2018virtual,xie2019unsupervised},
this loss is optimized \wrt\ $\hat{\mathcal{C}}^{\text{p}}_{n}, \hat{\mathcal{R}}^{\text{p}}_{n}$ (not \wrt\ $\mathbf{C}^{\text{p}}_{n}, \mathbf{R}^{\text{p}}_{n}$), \ie, $\bm{\theta}^{\text{b}}, \bm{\theta}^{\text{r-cnn}}, \bm{\theta}^{\text{cls}}, \bm{\theta}^{\text{reg}}$.
Then we can learn more noisy-robust proposal features and predictions.
\begin{equation}
\label{equ:l_cbpl}
\begin{aligned}
   &\mathrm{L}^{\text{cons}}\left(\mathbf{C}^{\text{p}}_{n}, \mathbf{R}^{\text{p}}_{n}, \hat{\mathcal{C}}^{\text{p}}_{n}, \hat{\mathcal{R}}^{\text{p}}_{n}\right) \\
   = & \ \lambda^{\text{cons-cls}} \mathrm{L}^{\text{cons-cls}}\left(\mathbf{C}^{\text{p}}_{n}, \hat{\mathcal{C}}^{\text{p}}_{n}\right) 
    + \lambda^{\text{cons-reg}} \mathrm{L}^{\text{cons-reg}}\left(\mathbf{R}^{\text{p}}_{n}, \hat{\mathcal{R}}^{\text{p}}_{n}\right).
\end{aligned}
\end{equation}

\section{Experiments}
\label{sec:exp}

In this section,
we will conduct thorough experiments to
analyze our proposal learning approach and its components
for semi-supervised object detection.

\subsection{Experimental Setup}
\label{sec:exp_setup}

\subsubsection{Dataset and evaluation metrics.}
We evaluate our approach on the challenging COCO dataset \cite{lin2014microsoft}.
The COCO dataset contains more than 200K images for 80 object classes.
Unlike many semi-supervised object detection works
conducting experiments on a simulated setting
by splitting a fully annotated dataset into labeled and unlabeled subsets,
we use all available labeled and unlabeled training data in COCO as our $\mathcal{D}^{\text{l}}$ and $\mathcal{D}^{\text{u}}$ respectively, following \cite{radosavovic2018data}.
More precisely, we use the COCO \texttt{train2017} set (118K images) as $\mathcal{D}^{\text{l}}$
and the COCO \texttt{unlabeled2017} set (123K images) as $\mathcal{D}^{\text{u}}$
to train object detectors.
In addition, we use the COCO \texttt{val2017} set (5K images) for validation and ablation studies,
and the COCO \texttt{test-dev2017} set (20K images) for testing.

We use the standard COCO criterion as our evaluation metrics,
including AP (averaged average precision over different IoU thresholds, the primary evaluation metric of COCO), AP$_{50}$ (average precision for IoU threshold 0.5), AP$_{75}$ (average precision for IoU threshold 0.75), AP$_{S}$ (AP for small objects), AP$_{M}$ (AP for medium objects), AP$_{L}$ (AP for large objects).

\begin{table*}[!tb]
\caption{Experimental results of different components of our proposal learning approach on the COCO \texttt{val2017} set.
ResNet-50 is chosen as our CNN backbone here.
No ``\checkmark'',
``$\mathrm{L}^{\text{self-loc}}$'',
``$\mathrm{L}^{\text{self-cont}}$'',
``$\mathrm{L}^{\text{cons-cls}}$'',
``$\mathrm{L}^{\text{cons-reg}}$'',
``PLLD'', and ``FSWA''
denote the fully-supervised baseline,
the proposal location loss in Eq.~(\ref{equ:l_location}),
the contrastive loss in Eq.~(\ref{equ:l_contrastive}),
the consistency loss for bounding box classification predictions in Eq.~(\ref{equ:l_cls}),
the consistency loss for bounding box regression predictions in Eq.~(\ref{equ:l_reg}),
Proposal Learning for Labeled Data,
and Fast Stochastic Weight Averaging, respectively.}
\begin{center}
\footnotesize
\resizebox{\linewidth}{!}{
\begin{tabular}{cccccc|cccccc}
  \toprule
  $\mathrm{L}^{\text{self-loc}}$ & $\mathrm{L}^{\text{self-cont}}$ & $\mathrm{L}^{\text{cons-cls}}$ & $\mathrm{L}^{\text{cons-reg}}$ & PLLD & FSWA & AP & AP$_{50}$ & AP$_{75}$ & AP$_{S}$ & AP$_{M}$ & AP$_{L}$\\
  \midrule
  & & & & & & 37.4 & 58.9 & 40.7 & 21.5 & 41.1 & 48.6 \\
  \checkmark & & & & & & 37.6$_{\uparrow 0.2}$ & 58.9$_{\uparrow 0.0}$ & 40.7$_{\uparrow 0.0}$ & 21.4$_{\downarrow 0.1}$ & 41.1$_{\uparrow 0.0}$ & 49.2$_{\uparrow 0.6}$ \\
  & \checkmark & & & & & 37.6$_{\uparrow 0.2}$ & 59.0$_{\uparrow 0.1}$ & 41.2$_{\uparrow 0.5}$ & 21.2$_{\downarrow 0.3}$ & 41.0$_{\downarrow 0.1}$ & 49.0$_{\uparrow 0.4}$ \\
  \checkmark & \checkmark & & & & & 37.7$_{\uparrow 0.3}$ & 59.2$_{\uparrow 0.3}$ & 40.6$_{\downarrow 0.1}$ & 21.8$_{\uparrow 0.3}$ & 41.3$_{\uparrow 0.2}$ & 49.1$_{\uparrow 0.5}$ \\
  & & \checkmark & & & & 37.8$_{\uparrow 0.4}$ & 59.2$_{\uparrow 0.3}$ & 41.0$_{\uparrow 0.3}$ & 21.6$_{\uparrow 0.1}$ & 41.2$_{\uparrow 0.1}$ & 50.1$_{\uparrow 1.5}$ \\
  & & & \checkmark & & & 37.6$_{\uparrow 0.2}$ & 59.0$_{\uparrow 0.1}$ & 40.8$_{\uparrow 0.1}$ & 21.0$_{\downarrow 0.5}$ & 41.3$_{\uparrow 0.2}$ & 49.2$_{\uparrow 0.6}$ \\
  & & \checkmark & \checkmark & & & 37.9$_{\uparrow 0.5}$ & 59.2$_{\uparrow 0.3}$ & 40.9$_{\uparrow 0.2}$ & 21.4$_{\downarrow 0.1}$ & 41.1$_{\uparrow 0.0}$ & 50.6$_{\uparrow 2.0}$ \\
  \checkmark & \checkmark & \checkmark & \checkmark & & & 38.0$_{\uparrow 0.6}$ & 59.2$_{\uparrow 0.3}$ & 41.1$_{\uparrow 0.4}$ & 21.6$_{\uparrow 0.1}$ & 41.5$_{\uparrow 0.4}$ & 50.4$_{\uparrow 1.8}$ \\
  \checkmark & \checkmark & \checkmark & \checkmark & \checkmark & & 38.1$_{\uparrow 0.7}$ & 59.3$_{\uparrow 0.4}$ & 41.2$_{\uparrow 0.5}$ & 21.7$_{\uparrow 0.2}$ & 41.2$_{\uparrow 0.1}$ & \textbf{50.7$\mathbf{_{\uparrow 2.1}}$} \\
  & & & & & \checkmark & 37.5$_{\uparrow 0.1}$ & 59.0$_{\uparrow 0.1}$ & 40.7$_{\uparrow 0.0}$ & 22.2$_{\uparrow 0.7}$ & 41.1$_{\uparrow 0.0}$ & 48.6$_{\uparrow 0.0}$ \\
  \checkmark & \checkmark & \checkmark & \checkmark & \checkmark & \checkmark & \textbf{38.4$\mathbf{_{\uparrow 1.0}}$} & \textbf{59.7$\mathbf{_{\uparrow 0.8}}$} & \textbf{41.7$\mathbf{_{\uparrow 1.0}}$} & \textbf{22.6$\mathbf{_{\uparrow 1.1}}$} & \textbf{41.8$\mathbf{_{\uparrow 0.7}}$} & 50.6$_{\uparrow 2.0}$ \\
  \bottomrule
\end{tabular}
}
\end{center}
\label{table:ablation}
\vspace{-0.3cm}
\end{table*}

\subsubsection{Implementation details.}
In our experiments,
we choose Faster R-CNN \cite{ren2017faster} with feature pyramid networks \cite{lin2017feature} and RoIAlign \cite{he2017mask} as our proposal-based object detectors,
which are the foundation of many recent state-of-the-art object detectors.
Different CNN backbones, including ResNet-50 \cite{he2016deep}, ResNet-101 \cite{he2016deep}, ResNeXt-101-32$\times$4d \cite{xie2017aggregated},
and ResNeXt-101-32$\times$4d with Deformable ConvNets \cite{zhu2019deformable} (ResNeXt-101-32$\times$4d+DCN), are chosen.

We train object detectors on 8 NVIDIA Tesla V100 GPUs for 24 epochs,
using stochastic gradient descent with momentum 0.9 and weight decay 0.0001.
During each training mini-batch,
we randomly sample one labeled image and one unlabeled image for each GPU,
and thus the effective mini-batch size is 16.
Learning rate is set to 0.01 and is divided by 10 at the 16$^{\text{th}}$ and 22$^{\text{nd}}$ epochs.
We use linear learning rate warm up to increase learning rate from 0.01/3 to 0.01 linearly in the first 500 training iterations.
In addition, object detectors are first trained only on labeled data using loss Eq.~(\ref{equ:l_labeled}) for 6 epochs.
We also use fast stochastic weight averaging for checkpoints from the last few epochs for higher performance, following \cite{athiwaratkun2018there}.

Loss weights $\lambda^{\text{self-loc}}, \lambda^{\text{self-cont}}, \lambda^{\text{cons-cls}}, \lambda^{\text{cons-reg}}$ are set to $0.25, 1, 1, 0.5$, respectively.
We add two types of noise, DropBlock \cite{ghiasi2018dropblock} with block size 2 and SpatialDropout \cite{tompson2015efficient} with dropout ratio 1/64, to proposal convolutional feature maps.
Other types of noise are also possible and we find that these two types of noise work well.
The number of noisy proposal features for each proposal is set to 4, \ie, $K=4$.
The temperature hyper-parameter $\tau$ in Eq.~(\ref{equ:l_contrastive}) is set to 0.1, following \cite{ye2019unsupervised}.
Images are resized so that the shorter side is 800 pixels with/without random horizontal flipping for training/testing.
Considering that most of the proposals mainly contain backgrounds,
we only choose the positive proposals for labeled data
and the proposals having maximum object score larger than 0.5 for unlabeled data
to compute proposal learning based losses,
which ensures networks focusing more on objects than backgrounds.

Our experiments are implemented based on the PyTorch \cite{paszke2019pytorch} deep learning framework
and the MMDetection \cite{chen2019mmdetection} toolbox.

\subsection{Ablation Studies}
\label{sec:ablation}

In this part,
we conduct elaborate experiments to analyze the influence of different components of our proposal learning approach,
including different losses defined in Eq.~(\ref{equ:l_location}), (\ref{equ:l_contrastive}), (\ref{equ:l_cls}), (\ref{equ:l_reg}), applying proposal learning for labeled data, and Fast Stochastic Weight Averaging (FSWA).
Without loss of generality, we only use ResNet-50 as our CNN backbone.
For the first two subparts, we only apply proposal learning to unlabeled data, \ie, using Eq.~(\ref{equ:l_all}).
For the first three subparts, we do not use FSWA.
Results are shown in Table~\ref{table:ablation},
where the first row reports results from fully-supervised baseline,
\ie, training object detectors only on labeled data using Eq.~(\ref{equ:l_labeled}).
For all experiments,
we fix the initial random seed during training to ensure
that the performance gains come from our approach instead of randomness.

\vspace{0.1cm}
\noindent\textbf{Self-supervised proposal learning.}
We first discuss the influence of our self-supervised proposal learning module.
As shown in Table~\ref{table:ablation},
compared to the fully-supervised baseline,
both the proposal location loss and the contrastive loss obtain higher AP (37.6\% \vs 37.4\% and 37.6\% \vs 37.4\% respectively).
The combination of these two losses,
which forms the whole self-supervised proposal learning module,
improves AP from 37.4\% to 37.7\%,
which confirms the effectiveness of the self-supervised proposal learning module.
The AP gains come from that
this module learns better proposal features from unlabeled data.

\vspace{0.1cm}
\noindent\textbf{Consistency-based proposal learning.}
We then discuss the influence of our consistency-based proposal learning module.
From Table~\ref{table:ablation},
we observe that applying consistency losses to both bounding box classification and regression predictions
obtains higher AP than the fully-supervised baseline
(37.8\% \vs 37.4\% and 37.6\% \vs 37.4\% respectively).
The combination of these two consistency losses,
which forms the whole consistency-based proposal learning module,
improves AP from 37.4\% to 37.9\%.
The AP gains come from that
this module learns better proposal features and predictions from unlabeled data.
In addition,
after combining the consistency-based proposal learning and self-supervised proposal learning modules,
\ie, using our whole proposal learning approach,
the AP is improved to 38.0\% further,
which shows the complementarity between our two modules.

\vspace{0.1cm}
\noindent\textbf{Proposal learning for labeled data.}
We also apply our proposal learning approach to labeled data,
\ie, using Eq.~(\ref{equ:l_all_new}).
From Table~\ref{table:ablation},
we see that proposal learning for labeled data boosts AP
from 38.0\% to 38.1\%.
This is because our proposal learning benefits from more training data.
The results also show that
our proposal learning can improve fully-supervised object detectors potentially,
but since we focus on semi-supervised object detection,
we would like to explore this in the future.

\vspace{0.1cm}
\noindent\textbf{Fast stochastic weight averaging.}
We finally apply FSWA to our approach
and obtain 38.4\% AP, as shown in Table~\ref{table:ablation}.
This result suggests that FSWA can also boost performance for semi-supervised object detection.
To perform a fair comparison, we also apply FSWA to the fully-supervised baseline.
FSWA only improves AP from 37.4\% to 37.5\%.
The results demonstrate FSWA gives more performance gains for our approach compared to the fully-supervised baseline.

According to these results,
in the rest of our paper,
we use all components of our proposal learning on both labeled and unlabeled data (\ie, using Eq.~(\ref{equ:l_all_new})),
and apply FSWA to our approach.

\begin{table}[!tb]
\caption{Experimental result comparisons among fully-supervised baselines (no ``\checkmark''), Data Distillation (DD) \cite{radosavovic2018data}, and our approach (Ours) on the COCO \texttt{test-dev2017} set.
Different CNN backbones are chosen.
Results of DD are reproduced by ourselves and are comparable with or even better than the results reported in the original DD paper.}
\begin{center}
\footnotesize
\resizebox{\linewidth}{!}{
\begin{tabular}{c|cc|cccccc}
  \toprule
  CNN backbone & DD & Ours & AP & AP$_{50}$ & AP$_{75}$ & AP$_{S}$ & AP$_{M}$ & AP$_{L}$\\
  \midrule
  \multirow{4}{*}{ResNet-50} & & & 37.7 & 59.6 & 40.8 & 21.6 & 40.6 & 47.2 \\
  & \checkmark & & 38.5 & 60.4 & 41.7 & 22.5 & 41.9 & 47.4 \\
  & & \checkmark & 38.6 & 60.2 & 41.9 & 21.9 & 41.4 & 48.9 \\
  & \checkmark & \checkmark & \textbf{39.6} & \textbf{61.5} & \textbf{43.2} & \textbf{22.9} & \textbf{42.9} & \textbf{49.8} \\
  \midrule
  \multirow{4}{*}{ResNet-101} & & & 39.6 & 61.2 & 43.2 & 22.2 & 43.0 & 50.3 \\
  & \checkmark & & 40.6 & 62.2 & 44.3 & 23.2 & 44.4 & 50.9 \\
  & & \checkmark & 40.4 & 61.8 & 44.2 & 22.6 & 43.6 & 51.6 \\
  & \checkmark & \checkmark & \textbf{41.3} & \textbf{63.1} & \textbf{45.3} & \textbf{23.4} & \textbf{45.0} & \textbf{52.7} \\
  \midrule
  \multirow{4}{*}{ResNeXt-101-32$\times$4d} & & & 40.7 & 62.3 & 44.3 & 23.2 & 43.9 & 51.6 \\
  & \checkmark & & 41.8 & 63.6 & 45.6 & 24.5 & 45.4 & 52.4 \\
  & & \checkmark & 41.5 & 63.2 & 45.4 & 23.9 & 44.8 & 52.8 \\
  & \checkmark & \checkmark & \textbf{42.8} & \textbf{64.4} & \textbf{46.9} & \textbf{24.9} & \textbf{46.4} & \textbf{54.4} \\
  \midrule
  \multirow{4}{*}{ResNeXt-101-32$\times$4d+DCN} & & & 44.1 & 66.0 & 48.2 & 25.7 & 47.3 & 56.3 \\
  & \checkmark & & 45.4 & 67.0 & 49.6 & 27.3 & 49.0 & 57.9 \\
  & & \checkmark & 45.1 & 66.8 & 49.2 & 26.4 & 48.3 & 57.6 \\
  & \checkmark & \checkmark & \textbf{46.2} & \textbf{67.7} & \textbf{50.4} & \textbf{27.6} & \textbf{49.6} & \textbf{59.1} \\
  \bottomrule
\end{tabular}
}
\end{center}
\label{table:test}
\vspace{-0.3cm}
\end{table}

\subsection{Main Results}
\label{sec:main_results}

We report the result comparisons among fully-supervised baselines,
Data Distillation (DD) \cite{radosavovic2018data},
and our approach in Table~\ref{table:test} on the COCO \texttt{test-dev2017} set.
As we can see, our approach obtains consistently better results compared to the fully-supervised baselines for different CNN backbones.
In addition, both DD and our approach obtain higher APs than the fully-supervised baselines,
which demonstrates that training object detectors on both labeled and unlabeled data outperforms training object detectors only on labeled data,
confirming the potentials of semi-supervised object detection.
Using our approach alone obtains comparable APs
compared to DD.

In particular, we also evaluate the efficiency of our method by combining with DD.
More specifically, we first train object detectors using our approach,
then follow DD to label unlabeled data,
and finally re-train object detectors using both fully-supervised loss and proposal learning losses.
The combination of our approach and DD obtains 39.6\% (ResNet-50), 41.3\% (ResNet-101), 42.8\% (ResNeXt-101-32$\times$4d), and 46.2\% (ResNeXt-101-32$\times$4d+DCN) APs,
which outperforms the fully-supervised baselines by about 2.0\% on average
and DD alone by 0.9\% on average.
The results demonstrate that our approach and DD are complementary to some extent.

\section{Conclusion}
\label{sec:con}

In this paper,
we focus on semi-supervised object detection for proposal-based object detectors (a.k.a. two-stage object detectors).
To this end,
we present a proposal learning approach,
which consists of a self-supervised proposal learning module
and a consistency-based proposal learning module,
to learn proposal features and predictions from both labeled and unlabeled data.
The self-supervised proposal learning module
learns context-aware and noise-robust proposal features
by a proposal location loss and a contrastive loss respectively.
The consistency-based proposal learning module
learns noise-robust proposal features and predictions
by consistency losses for both bounding box classification and regression predictions.
Experimental results show that
our approach outperforms fully-supervised baselines consistently.
It is also worth mentioning that
we can further boost detection performance by combining our approach and data distillation.
In the future,
we will explore more self-supervised learning and semi-supervised learning ways for semi-supervised object detection,
and explore how to apply our approach to semi-supervised instance segmentation.

{\small
\bibliographystyle{ieee_fullname}
\bibliography{egbib}
}

\end{document}